%% file: main.tex
\documentclass[runningheads]{llncs}
\usepackage[T1]{fontenc}
\usepackage{url}
\usepackage{graphicx}
\usepackage{amssymb}
\usepackage{amsmath}
\usepackage{authblk}
\usepackage{algorithm}
\usepackage{algorithmic}
\usepackage{microtype}
\usepackage{tikz}
\usepackage{pgf}
\usepackage{lipsum}
\usepackage{verbatim}
\usetikzlibrary{positioning, arrows.meta, shapes.geometric}
\usepackage[hidelinks]{hyperref}
\usepackage{xcolor}

\usepackage{color}

\begin{document}
\title{Deep Reinforcement Learning Based Navigation with Macro Actions and Topological Maps}

\titlerunning{Deep RL based Navigation with Macro Actions and Topological Maps}
%
\author{Simon Hakenes \and Tobias Glasmachers}
\authorrunning{S. Hakenes and T. Glasmachers}
%
\institute{Institute for Neural Computation, Faculty of Computer Science, Ruhr~University~Bochum, Germany \\ \email{simon.hakenes@ini.rub.de, tobias.glasmachers@ini.rub.de}}
\maketitle              
\begin{abstract} This paper addresses the challenge of navigation in large, visually complex environments with sparse rewards. We propose a method that uses object-oriented macro actions grounded in a topological map, allowing a simple Deep Q-Network (DQN) to learn effective navigation policies. The agent builds a map by detecting objects from RGBD input and selecting discrete macro actions that correspond to navigating to these objects. This abstraction drastically reduces the complexity of the underlying reinforcement learning problem and enables generalization to unseen environments. We evaluate our approach in a photorealistic 3D simulation and show that it significantly outperforms a random baseline under both immediate and terminal reward conditions. Our results demonstrate that topological structure and macro-level abstraction can enable sample-efficient learning even from pixel data.
\end{abstract}
\section{Introduction}
There are many practical situations where an agent must navigate extensive environments and interact with different objects following a strict sequence. 
An example of such a situation is an industrial manufacturing process where a product needs to be picked up in a warehouse, transported to several machines and dropped off somewhere else. Another one would be a video game, where the map is unknown beforehand, a level must be explored efficiently, a key must be found and in the end the player must remember the position of the door it unlocks. 

Navigation in such extensive environments is still an unsolved problem for standard reinforcement learning (RL) methods, especially when rewards are sparse, the agent is controlled with elementary actions (like single steps or rotations), and training relies on visual data. In such cases, agents struggle with huge state spaces and inefficient exploration, making effective learning practically impossible.

This is at least partly because standard RL methods lack basic world knowledge, such as scene understanding and the concept of 3D space, including what it means to move forward or turn. 
Incorporating these concepts into the algorithm seems like a natural improvement, in contrast to learning them by relying solely on the Markov Decision Process~\cite{Sutton1999-yp}. 
Humans, in contrast, rely on their pre-structured brain as well as prior experience to learn similar concepts instead of starting from scratch. 
Concepts of objects, actions, numbers, and space are understood from an early age, enabling us to learn navigation fast~\cite{Spelke2007-fl}. 
This understanding naturally forms an internal representation of the scene and even a map of the environment, which allows us to plan routes, remember important locations, and navigate efficiently, even in unfamiliar surroundings~\cite{Parra-Barrero2023-xm}.

Just as humans rely on innate spatial knowledge, an artificial agent can benefit from structured world knowledge to navigate efficiently. 
Technically, all essential components for building a navigation agent exist already. Simultaneous Localization and Mapping (SLAM)~\cite{campos2021orb,Dissanayake2001-dm,Engel2014-qy} provides spatial awareness, segmentation and object detection enable scene understanding, graph representations facilitate efficient path planning, and reinforcement learning offers a framework for decision-making. 
However, these components are typically used in isolation and are rarely combined into a unified system tailored for navigation in large, complex environments.

A key enabler for an efficient navigating agent is a good map representation. In this work, we use topological maps, which are also found in the mammalian brain~\cite{Parra-Barrero2023-xm}, as they provide an efficient structure for planning and navigation. The main advantage of topological maps -- their flexibility and ability to grow to arbitrary size -- is also their biggest disadvantage when it comes to using them in a deep RL setting. Using a graph as the state and/or action space of an RL algorithm requires dynamically changing the sizes of state and action spaces.

\begin{figure}[!htb]
    \centering
    \includegraphics[width=0.6\linewidth]{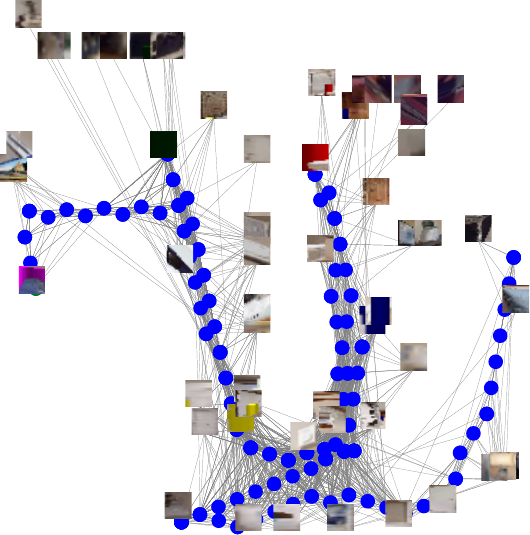}
    \caption{Example of a topological map built by the agent. Object nodes are visualized using one representative image per object, while waypoints are shown as blue dots. Edges indicate navigable connections. See Section~\ref{sec:map} for a more detailed description.}
    \label{fig:map}
\end{figure}

In this paper we consider actions specific to a topological map. An action is defined as navigating to a specific node on the map. Since this usually involves many elementary control commands, we refer to it as a macro action. As the agent explores, the number of nodes, and thus the number of available actions, continuously increases (Fig.~\ref{fig:map}). 

A major challenge is integrating these elements while handling dynamically growing state and action spaces. Standard RL methods assume fixed action spaces, making them unsuitable for topological maps that continuously expand as the agent explores. Furthermore, sparse rewards and inefficient exploration make it nearly impossible for traditional deep RL approaches to learn effective policies.

The main contributions of this paper are as follows. 
First, we integrate object-oriented topological maps with a novel Deep Q-Network (DQN) architecture, allowing for more structured and efficient navigation. Second, we establish a direct connection between the topological map and the DQN, improving exploration and navigation performance in complex environments. Finally, we streamline the overall process to reduce the computational burden on the RL algorithm, enabling the DQN to solve an otherwise intractable navigation task. 
We showcase our approach in a photorealistic, multi-room simulation environment with diverse target objects (Fig.~\ref{fig:env_screenshots}).

\section{Related Work}

A naive approach to navigation in RL is to use only elementary actions in a partially observable Markov decision process (POMDP), combined with a feed-forward neural network. The agent receives limited observations (e.g. first-person RGB images) and must learn to navigate through trial and error without any memory or map. That is shown in \cite{Mirowski2017-im}, however generalization to unseen environments is questionable. 

Temporal abstraction through the options framework~\cite{Sutton1999-yp} allows agents to learn sequences of primitive actions, called options, to simplify decision-making. While related work~\cite{Ramesh2019-hr,Stolle2002-zi} applies this approach to navigation, the learned options often lack generalization. In our work, we refer to similar abstractions as macro actions, but they are designed around object-based targets and do not need to be learned.

Metric maps are a common and promising way to represent spatial environments, capturing accurate geometry and supporting precise localization and planning. 
They can be implemented in a differentiable way using spatial memory structures with learnable and fully differentiable read and write operations, enabling end-to-end training~\cite{Parisotto2018-an,Zhang2017-ay}. 
However, such approaches typically lack positional readout, are limited in size, and cannot grow with the environment. 
Moreover, letting the agent learn what to store is unnecessary, as this kind of spatial and semantic information can be provided directly.

An example of a non-differentiable map can be found in \cite{Wani2020-km} on which the task in this work is based. 
However, the reward that is based on the distance decrease to the next goal makes the task in principle solvable without any map. 
Unlike their setup, we assume that the order of the objects is fixed across episodes but completely unknown to the agent.

In any case, representing the environment as a graph is a powerful approach, because it gives more flexibility and also path planning becomes straightforward. 
Some previous work used topological maps, but their systems are not trained with RL \cite{Savinov2018-pg}. 

A closely related work is TopoNav~\cite{Hossain2024-sv}, developed independently and published shortly after our preliminary version~\cite{Hakenes2023-gj}. 
They apply hierarchical deep reinforcement learning to topological navigation. 
Their method combines a hierarchical DQN architecture~\cite{Kulkarni2016-eo} with dynamic topological map construction, using a meta-controller to select subgoals and sub-controllers to reach them. 
While this is conceptually similar to our use of topological maps and high-level decisions, there are several key differences.

First, their subgoals are learned implicitly through feature detection and strategic selection based on novelty and goal-directedness, whereas we define macro actions explicitly as object-based navigation targets. This removes the need for learned subgoal discovery and introduces strong priors that simplify learning. Second, their architecture requires separate networks for meta and sub-controllers, as well as carefully tuned intrinsic reward shaping for exploration. In contrast, we use a single DQN and focus on simplifying the RL problem through abstraction rather than architectural complexity. Finally, TopoNav evaluates real-world robotics scenarios, while our work emphasizes algorithmic simplicity and analysis under idealized assumptions (e.g., perfect object detection and SLAM) to isolate the learning dynamics.

Our approach can be seen as a more minimalistic, object-centric alternative to TopoNav that avoids handcrafted reward shaping and complex multi-agent training schemes, focusing instead on how structure and abstraction alone can make otherwise intractable RL problems solvable.

Finally, there is prior work on dynamically growing action spaces.
The authors of \cite{He2016-ui} address the problem of unbounded action spaces in text-based decision-making problems with a natural language action spaces. 
They introduce the Deep Reinforcement Relevance Network (DRRN) that has two inputs. One processes the state embedding while the other processes the action embedding. They are combined with a pairwise interaction function (e.g. an inner product), outputting a single value.

\section{Method}
The main idea of this work is to create macro actions that enable the agent to efficiently navigate an environment and learn how to solve problems there.
A macro action navigates to a specific object that is already stored on the map. 
If the agent visits objects one after another, we assume that it can navigate between them trivially without problems, forming a topological map of the environment.

A macro action is related to an option in the options framework \cite{Sutton1999-yp}. However, our macro actions do not require learning because using a map is in itself not a hard but rather a mechanical task. Of course, doing so based on vision can be highly non-trivial. That is the subject of SLAM algorithms, which are outside of the scope of this paper. In contrast, we focus on the creation of an object-centric topological map, and on learning to solve tasks based on that map. In our map design (explained below) the low-level controller is applied only to rather short distances, making its task tractable.

\begin{figure}
    \centering
    \begin{minipage}{0.16\linewidth}
        \includegraphics[width=\linewidth]{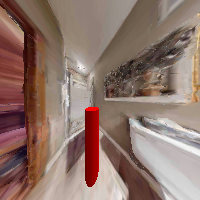}
        \centering (a)
    \end{minipage}\hfill
    \begin{minipage}{0.16\linewidth}
        \includegraphics[width=\linewidth]{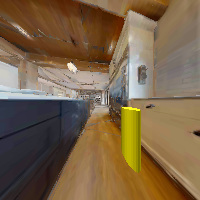}
        \centering (b)
    \end{minipage}\hfill
    \begin{minipage}{0.16\linewidth}
        \includegraphics[width=\linewidth]{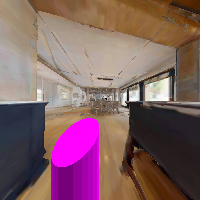}
        \centering (c)
    \end{minipage}\hfill
    \begin{minipage}{0.16\linewidth}
        \includegraphics[width=\linewidth]{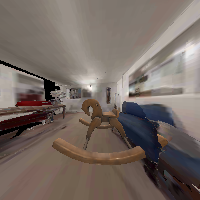}
        \centering (d)
    \end{minipage}\hfill
    \begin{minipage}{0.16\linewidth}
        \includegraphics[width=\linewidth]{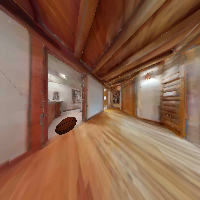}
        \centering (e)
    \end{minipage}\hfill
    \begin{minipage}{0.16\linewidth}
        \includegraphics[width=\linewidth]{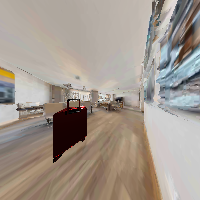}
        \centering (f)
    \end{minipage}
    \caption{Screenshots of the environments with different target objects. In (a)–(c), the targets are color-coded cylinders, while in (d) the target is a rocking horse, in (e) a basketball, and in (f) a suitcase. Each environment consists of multiple rooms.}
    \label{fig:env_screenshots}
\end{figure}

In the following subsection we explain how the topological map is created while the agent explores the photorealistic environment (see Fig.~\ref{fig:env_screenshots} for exemplary screenshots). 
How the agent navigates in the environment is outlined in Section~\ref{sec:navigation}.
The integration of the growing and unbounded map makes it necessary to rethink action and state spaces, which we will do in Section~\ref{sec:actionspace}. Lastly, we explain how our novel DQN can train under these conditions. 

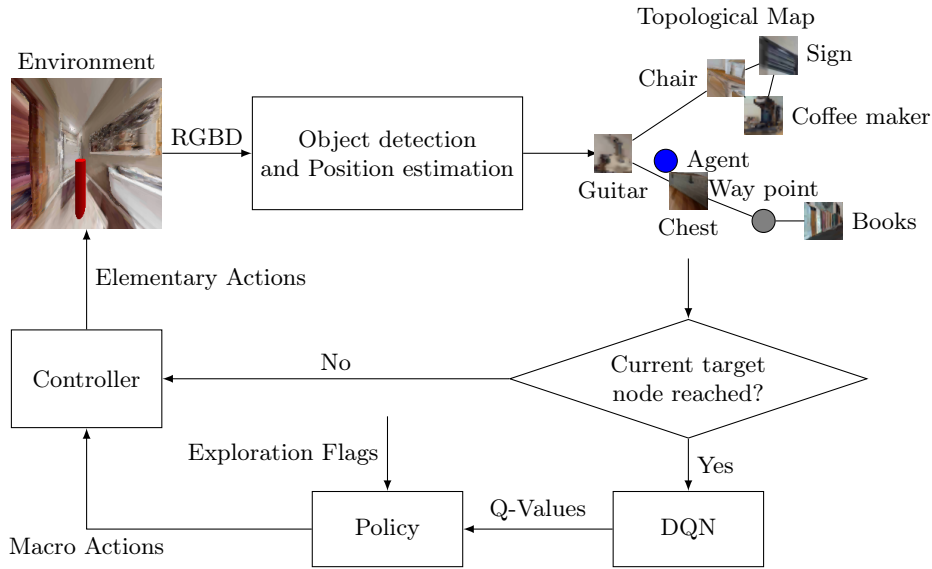
\begin{figure}
    \centering
    \input{fig_overview}
    \caption{Overview of the system.}
    \label{fig:overview}
\end{figure}

\subsection{Map Representation}\label{sec:map}
While moving through the environment, the agent perceives RGB frames with depth information (RGBD). This is a standard requirement of many state-of-the-art SLAM algorithms~\cite{campos2021orb,Dissanayake2001-dm,Engel2014-qy}. Depth information is easy to obtain in simulations, while it requires stereo vision or time-of-flight cameras in the real world.

The agent detects all objects in each frame using ground truth data provided by the environment. While this can be replaced by an object detection module in future work, it is not the focus of this paper.
From the agent's position and viewing angle, the depth information, and the position of the object in the frame, the position in 3D space of the object can be estimated. 
For every object a node will be created on the map. 
We store the corresponding pixels inside the bounding box (rescaled to $16 \times 16$~px), the 3D coordinates and a flag whether the object is already explored inside of the node structure. Moving to unexplored nodes often brings new locations into view, which results in a natural growth process of the map until the whole environment is explored.

The nodes are marked as \emph{explored} (in contrast to \emph{seen from a distance}) as soon as the agent is close to the node, greatly improving the exploration behavior, since we can easily give higher priority to unexplored nodes during exploration.
Edges are added to the graph when a line-of-sight between two objects is detected or the agent moves from one object to another, creating a navigable graph of the environment (see Fig.~\ref{fig:map} and top part of Fig.~\ref{fig:overview}).

\begin{figure}
    \centering
    \includegraphics[width=\linewidth]{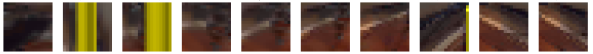}
    \caption{A set of input patches of an object. Note how in some perspectives the object is occluded by the yellow target cylinders.}
    \label{fig:input_patches}
\end{figure}

Since the agent observes the same object from different angles and distances, often with partial occlusions, each object node stores multiple image patches. Figure~\ref{fig:input_patches} shows several views of the same object, some of which are partially occluded by the yellow target cylinder. While a single image might be ambiguous or misleading, aggregating multiple views allows the agent to form a more robust representation of the object. 
This design helps prevent false associations (e.g., confusing non-target objects for targets) and improves the stability of the learned value function during training.

Intermediate waypoints are added to keep the distances between nodes reasonably small. They serve as additional navigation aids, making paths between nodes easy to traverse. These waypoints can be positions the agent has already visited, ensuring that they are navigable.

\subsection{Navigating the Environment}\label{sec:navigation}
With the topological map formed, the agent can navigate the environment systematically. 
If there is no current target node for the agent or if it is reached, the DQN is rating all nodes currently stored on the map with Q-Values (Fig.~\ref{fig:overview}). This step is further detailed in Section~\ref{sec:dqn}. Of course, navigation happens already while the map is still expanding due to exploration.

The policy picks the next macro action, i.e., the next target node. 
For exploratove actions it is guided by the exploration flags, so that currently unexplored nodes can be prioritized over already explored ones.

Executing a macro action usually entails determining and following a potentially long sequence of waypoints. The shortest path through the graph to the target node is calculated with the well known A* algorithm. For reaching the next waypoint, the low-level controller is responsible for executing a sequence of elementary actions. It implements a simple aiming behavior, turning the agent so that it faces the next node and moving forward until it is reached.

\subsection{State and Action Spaces}\label{sec:actionspace}
Defining the state and action spaces is not straightforward as they are intertwined. The state space must incorporate the map as well as a progress indicator. The action space consists of all the navigable objects on the map. Both spaces are growing dynamically and hence depend on a step counter $t$. In this section, we formalize the representation of both spaces. 

The map is represented by a graph $\mathcal{G}_t = (\mathcal{V}_t, \mathcal{E}_t)$ where nodes $v \in \mathcal{V}_t$ correspond to places in the environment. 
Edges $e \in \mathcal{E}_t$ indicate trivially navigable connections. 
Each node can either be a place in the environment, that is navigable and can help to make more navigable connections, or it represents an object. 
In the latter case, the node is associated with a set of feature vectors $\mathcal{F}_v$ which contain the aforementioned RGB pixels of the object, so $\mathcal{F}_v \subseteq \mathbb{R}^{16 \times 16 \times 3}$. 

Since the agent may decide to move to any node, the current action space equals the set of the nodes on the map, $A_t = \mathcal{V}_t$, where $t$ is the current time-step.

For the task of visiting objects in the environment in the right sequence (think of sequential production steps as an example), the state depends not only on the current target but also on how many previous targets have already been visited. 
To represent this progress, we use a one-hot encoded vector $x_t \in \{0,1\}^{N_T}$, where $N_T$ is the total number of targets. 
The active entry in $x_t$ indicates which target the agent is currently expected to find.

\begin{equation}
    x_t[i] =
    \begin{cases}
    1, & \text{if target } i \text{ is the current target} \\
    0, & \text{otherwise}
    \end{cases}
\end{equation}
where $[i]$ denotes the index access operator.

The state space can then be defined as the set of all feature vectors, i.e. $\mathcal{S}_t = \{\mathcal{F}_v \mid v \in \mathcal{V}_t\} \times [N]$, where $[N]$ denotes the $N$ possible values of $x_t$.

\subsection{Modified DQN Architecture}\label{sec:dqn}
Each node associated with an object on the map is a feasible action. 
Hence, as the map grows, the agent has more possible actions available, which makes the action space grow. That's problematic since in most standard (deep) RL methods the number of possible actions is hard-coded in the number of output neurons of the neural network, which cannot represent a changing action space.
Therefore, it is necessary to divert from the usual DQN algorithm, taking inspiration from~\cite{He2016-ui} who applied a DQN to a question answering problems with varying numbers of answers.  
To choose the next macro-action (i.e. an object) the DQN needs to assign a value to each action.

The DQN must now approximate the optimal action value function $Q^\ast (f_t, x_t)$ for all $f_t \in \mathcal{V}_t$. 
Since the number of actions can increase every time step while the network architecture shall remain fixed, the DQN computes them separately. 
Hence, our DQN has a single output neuron, while the action for which the value is to be computed becomes a second input, next to the observation.

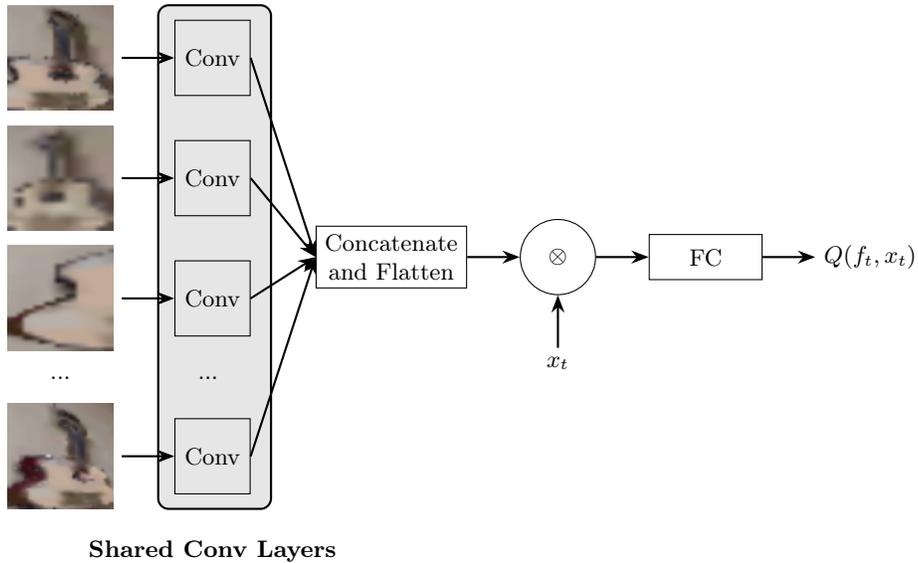
\begin{figure}
    \centering
    \input{network}
    \caption{Network architecture, where $\otimes$ denotes an outer product. 
             The input image represents a sample RGB observation fed into the network.
             The convolutional layers share the same weights.}
    \label{fig:network}
\end{figure}

The inputs to the neural network consists of an action representation, specifically $N_i$ random images $f_t$ corresponding to one node (or object), as well as the progress vector $x_t$. 
We handle them with a special architecture. 
Since the first input consists of images we apply a standard CNN approach. 
As usual, we think of the convolution layers as generic feature extractors, while the fully connected layers implement task-specific logic. 
All $N_i$ convolutional branches share their weights to accelerate learning of visual features. 
Since different actions require different logic, we merge the input streams by forming the outer product between the (flattened) output of the feature extractor and the vector $x_t$. 
The architecture is illustrated in Fig.~\ref{fig:network}.

The policy consists of two parts. 
One is targeted at exploration, the other one is greedy. 
The exploration part is chosen with probability $\epsilon$. 
It is based on a Boltzmann q-Policy \cite{Wiering1999-sv} with a temperature variable $T$ controling the greediness of the exploration. 
The probability $P(f|S_t)$ for selecting an action $f$ given state $S_t$ and Q-values $Q(f, x_t)$ for all $f \in \mathcal{V}_t$ is computed as follows:
\begin{equation}
    P(f|S_t) = \frac{\exp\left({ \frac{Q(f, x_t)+q}{T}}\right)} {\sum_{f \in \mathcal{V}_t} \exp\left({ \frac{Q(f, x_t)+q}{T}}\right)}
\end{equation}
with $q > 0$ for every unexplored node increasing the probability of exploring unexplored nodes over already explored ones. 

The greedy part is active with probability $1-\epsilon$. It selects the action with the highest q-value. If that action was already chosen previously (resulting in a no-operation), it instead selects the action with the next-highest Q-value. 

It is a known problem in deep RL that training of convolutional layers is difficult with the only learning signal being the sparse reward \cite{Hakenes2019-tg}. A further difficulty for visual learning is that objects can be (partially) occluded.  Therefore, we feed multiple images of the same object to the very same convolutional layers, resulting in more gradients from one backpropagation step greatly improving training speed and stability.

\section{Experiments and Results}
\subsection{Setup and Task Description}
For experiments we made use of the photorealistic Habitat Matterport 3D dataset~\cite{Savva2019-hh,Szot2021-ti}. 
Exemplary screenshots can be found in Fig.~\ref{fig:env_screenshots}.
The task for this project is to find a sequence of up to three target objects in an indoor scene consisting of multiple rooms with hundreds of objects. 
The positions of the target objects and the starting position of the agent vary between episodes but the order of the targets, which defines the type of task, remains the same.

There are two types of target objects: First, color-coded cylinders that are deliberately designed to be easily distinguishable, serving as a sanity check for the navigation system. Second, photo-realistic objects that naturally blend into the visual complexity of the environment. These require the agent to perform non-trivial visual processing to recognize them reliably, especially under varying viewpoints, lighting conditions, and partial occlusions. 
This setup ensures that the agent is capable of extracting visual features beyond simple color cues.

There are two different reward mechanisms: The first one provides an immediate reward of +1 each time the current target is found. 
The second one only gives a delayed terminal reward of +1 when \textit{all} targets are found. This setup introduces a significantly harder credit assignment problem, as the agent must execute long sequences of elementary actions without receiving intermediate feedback.

The agent has no prior knowledge of the goal and relies solely on the reward signal, which is extremely sparse relative to the elementary actions. 
Reaching a single target may require hundreds of elementary actions, making learning through random exploration nearly impossible.

Experiments are conducted in 100 different indoor scenes, each consisting of multiple rooms and different kinds of objects and varying target positions. 

We compare our trained agent against a random baseline that selects macro actions uniformly from the set of currently available objects. Both agents start with an empty map in environments with randomly placed targets. As they move, they perceive the scene via RGB frames, detect objects (using ground truth), and incrementally build a topological map.

The random agent selects a random object as its next macro action, while the trained agent uses its policy to either explore new areas or exploit existing knowledge by navigating to the current target. As both agents traverse the environment, their maps grow with each new observation. In both cases, the low-level controller (see Sec.~\ref{sec:navigation}) computes the shortest path to the selected node and executes the corresponding sequence of elementary actions. 

To select the next target node, the trained agent computes Q-values for all nodes on the map. For each node, 10 image patches are passed through the neural network, producing a Q-value estimate. The policy then selects a node either greedily or via Boltzmann q-Policy for exploration (Sec.~\ref{sec:dqn}).

\subsection{Results}
\begin{figure}
    \centering
    \input{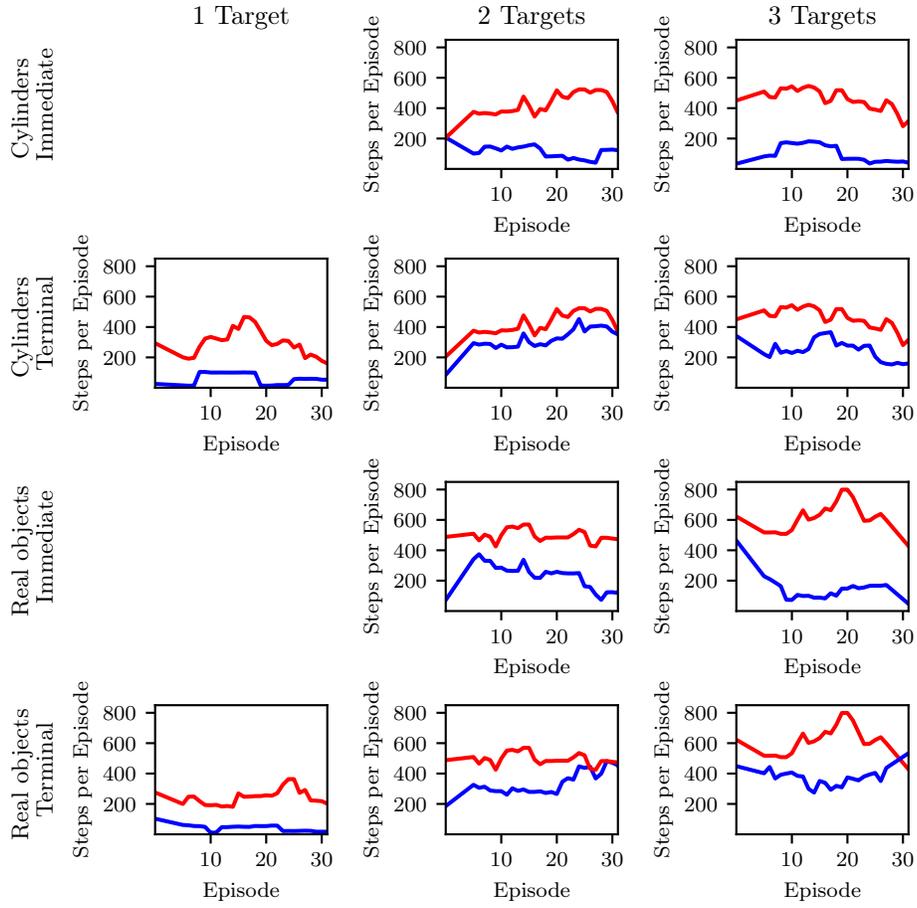}
    \caption{Plots showing steps per episode for 1, 2, and 3 targets with immediate and terminal reward.
    Red graphs depict random baseline (random macro actions) and blue graphs depict our trained agent. 
    Steps per episode are elementary actions.}
    \label{fig:results}
\end{figure}

Across all conditions, the trained agent consistently requires fewer steps per episode than the random baseline, demonstrating more efficient navigation (see Fig.~\ref{fig:results}). This performance gap is especially pronounced in the terminal reward setting, where sparse feedback makes learning more difficult. The benefit of training is also visible with real objects, where the agent must generalize over occlusions and viewpoint changes.
These results confirm that the agent learns to build and exploit the topological map effectively and can navigate complex environments with significantly improved efficiency over a random policy.

Note that the initial warm-up phase, during which the agent acts randomly to populate the replay buffer, is not shown in the plots. 
Since no steep performance increase is visible in the shown episodes, this suggests that the agent learned an effective policy within the first few training episodes. 
This rapid convergence highlights the benefits of our structured setup and macro-action abstraction.

\subsection{Discussion}
Our results show that the proposed method enables efficient learning and reliable navigation, even though we use a basic Deep Q-Network (DQN) architecture, which is generally considered outdated and ill-suited for complex navigation tasks. 
It is particularly noteworthy that the agent achieves good performance after only a few training episodes, despite the use of raw pixel input, which is traditionally difficult to learn from using reinforcement learning alone~\cite{Hakenes2019-tg}.

However, this rapid policy improvement should not be confused with low computational cost. Training requires significant wall time due to the large number of elementary actions executed between macro actions, leading to long episode durations despite few RL updates. Additionally, the method has high memory demands, driven by large batch sizes for stable training, the high number of objects on the map and the need to store multiple image patches per object node.

Another open challenge is the reliance on idealized low-level navigation. 
In this work, navigation between macro targets is executed via shortest-path planning with perfect localization and object detection. 
In future work, this component could be replaced or augmented by a SLAM~\cite{campos2021orb,Dissanayake2001-dm,Engel2014-qy}, which would allow for more realistic and robust integration of the agent’s perception and mapping capabilities.

One could argue that providing the agent with the one-hot encoded progress vector $x_t$ undermines the challenge of the terminal reward setting, as it effectively signals when an intermediate goal has been reached. However, this information could also be inferred from the sequence of past actions or observations. Our formulation simply encodes this information explicitly to reduce complexity and focus the learning on the navigation policy itself.

\section{Conclusion and Future Work}
This work demonstrates that macro actions based on topological maps can drastically reduce the complexity of navigation tasks in sparse-reward environments. We show that even a simple DQN can learn to solve a challenging 3D navigation task from raw pixel input despite the growing state and action space as the environment is explored.

A key strength of our approach is that it handles this dynamic action space without requiring a complex architecture. 
By abstracting the environment into object-based macro actions and topological structure, the agent learns a navigation policy that generalizes well to unseen environments, without overfitting to specific layouts or target configurations.

At the same time, the system remains simple: objects are detected, placed on a map, and used as navigation targets. The agent reasons over this structure directly, enabling effective exploration and goal-directed behavior based purely on visual input and sparse rewards.

That said, the current method assumes perfect object detection and localization, and depends on handcrafted heuristics for map construction. 
Training is also time-consuming due to the large number of elementary actions between macro steps and the need to process multiple views per object.

Future work should explore more realistic perception and mapping components (e.g., SLAM), remove the hand-crafted progress tracking vector, and experiment with alternative network architectures for improved feature learning and policy efficiency. 
Additionally, the current agent does not reason about the structure of the topological map itself. It treats all nodes independently without considering its own location, the spatial relations between objects, or the overall connectivity. 
Leveraging this structural information more explicitly could lead to more informed action selection and more efficient navigation.


%
%
%
\bibliographystyle{splncs04}
\bibliography{references}
\end{document}

%% file: fig_overview.tex
\begin{tikzpicture}[>=latex,scale=1]
    \node[inner sep=0] (image) at (0,0) {\includegraphics[width=2cm]{images/env1.png}};
    \node[above] at (0,1) {Environment};
    \node[draw, rectangle, align=center, minimum width=3cm, minimum height=1.5cm] (box) at (4,0) {Object detection \\ and Position estimation};
    
    \draw[->] (image.east) -- node[above] {RGBD} (box.west);

    \draw[->] (box.east) -- ++(1,0);

    \node[align=center] (map) at (8.5, 1.8) {Topological Map};
    
    \node[draw, circle, fill=blue] (player) at (7.7,-0.1) {};
    \node[inner sep=0] (guitar) at (7,0) {\includegraphics[width=0.5cm]{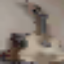}};
    \node[inner sep=0] (chair) at (8.5,1) {\includegraphics[width=0.5cm]{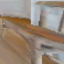}};
    \node[inner sep=0] (chest) at (8,-0.5) {\includegraphics[width=0.5cm]{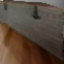}};
    \node[inner sep=0] (coffee_maker) at (9,0.5) {\includegraphics[width=0.5cm]{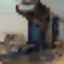}};
    \node[inner sep=0] (sign) at (9.2,1.3) {\includegraphics[width=0.5cm]{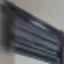}};
    \node[inner sep=0] (books) at (9.8,-0.9) {\includegraphics[width=0.5cm]{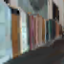}};
    \node[draw, circle, fill=gray] (waypoint) at (9.0,-0.9) {};

    \draw[-] (guitar) -- (chair);
    \draw[-] (guitar) -- (chest);
    \draw[-] (chest) -- (waypoint);
    \draw[-] (books) -- (waypoint);
    \draw[-] (chair) -- (sign);
    \draw[-] (chair) -- (coffee_maker);
    \draw[-] (sign) -- (coffee_maker);

    \node[right] at (player.east) {Agent};
    \node[above] at (waypoint.north) {Way point};
    \node[below] at (guitar.south) {Guitar};
    \node[left] at (chair.west) {Chair};
    \node[below] at (chest.south) {Chest};
    \node[right] at (coffee_maker.east) {Coffee maker};
    \node[right] at (sign.east) {Sign};
    \node[right] at (books.east) {Books};

    \node[draw, diamond, align=center, aspect=3] (if) at (8,-3) {Current target\\node reached?};
    
    \draw[->] (8,-1.4) -- (if.north);
    
    \node[draw, rectangle, align=center,minimum width=2cm, minimum height=1.3cm] (controller) at (0, -3) {Controller};
        
    \draw[->] (controller.north) -- node[right] {Elementary Actions} (image.south);
    \draw[->] (if.west) -- node[above] {No} (controller.east);
    
    \node[draw, rectangle, align=center, minimum width=2cm, minimum height=1cm] (policy) at (4, -5) {Policy};
    
    \node[draw, rectangle, align=center, minimum width=2cm, minimum height=1cm] (dqn) at(policy -| if) {DQN};
    \draw[->] (if.south) -- node[right] {Yes} (dqn.north);
    
	\draw[->] (4,-3.5) -- node[left] {Exploration Flags} (policy.north);
    \draw[->] (dqn.west) -- node [above]{Q-Values} (policy.east);

    \draw[->] (policy.west) -| node[below]{Macro Actions} (controller.south);
\end{tikzpicture}

%% file: network.tex
\begin{tikzpicture}[>=Stealth, transform shape, node distance=0.7cm]
      \tikzstyle{every node}=[font=\small]

      \node (input1) [align=center] at (0, 0)    {\includegraphics[width=1.4cm]{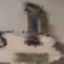}};
      \node (input2) [align=center] at (0, -1.6) {\includegraphics[width=1.4cm]{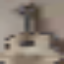}};
      \node (input3) [align=center] at (0, -3.2) {\includegraphics[width=1.4cm]{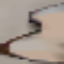}};
      \node (input4) [align=center, minimum width=1cm, minimum height=1cm] at (0, -4.25) {...};
      \node (input5) [align=center] at (0, -5.3) {\includegraphics[width=1.4cm]{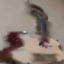}};

        \begin{scope}
        \fill[gray!20, rounded corners] (1.3,0.7) rectangle (2.8,-6);
        \draw[thick, rounded corners] (1.3,0.7) rectangle (2.8,-6);
      \end{scope}

      \node (conv1)  [draw, rectangle, align=center, minimum width=1cm, minimum height=1cm, right=of input1] {Conv};
      \node (conv2)  [draw, rectangle, align=center, minimum width=1cm, minimum height=1cm, right=of input2] {Conv};
      \node (conv3)  [draw, rectangle, align=center, minimum width=1cm, minimum height=1cm, right=of input3] {Conv};
      \node (conv4)  [align=center,                  minimum width=1cm, minimum height=1cm, right=of input4]{\hspace{5mm}...};
      \node (conv5)  [draw, rectangle, align=center, minimum width=1cm, minimum height=1cm, right=of input5] {Conv};

      \node [below=of conv5, yshift=0.2cm] {\textbf{Shared Conv Layers}};
      \node (flatten)  [draw, rectangle, align=center, minimum width=2cm, minimum height=0.6cm] at (4.4,-2.65) {Concatenate\\and Flatten};

      \node (outer)  [draw, circle, right=of flatten, minimum size=1cm] {$\otimes$};

      \node (input6) [below=of outer, align=center, text width=1.5cm] {$x_t$};

      \node (fully)  [draw, rectangle, right=of outer, minimum width=1.5cm, minimum height=0.6cm] {FC};

      \node (output) [right=of fully, minimum size=0.75cm] {$Q (f_t, x_t)$};

      \draw[->, thick] (input1.east) -- (conv1.west);
      \draw[->, thick] (input2.east) -- (conv2.west);
      \draw[->, thick] (input3.east) -- (conv3.west);
      \draw[->, thick] (input5.east) -- (conv5.west);

      \draw[->, thick] (conv1.east) -- (flatten.west) ;
      \draw[->, thick] (conv2.east) -- (flatten.west);
      \draw[->, thick] (conv3.east) -- (flatten.west);
      \draw[->, thick] (conv5.east) -- (flatten.west);

      \draw[->, thick] (flatten.east) -- (outer.west);
      \draw[->, thick] (input6.north) -- (outer.south);
      \draw[->, thick] (outer.east) -- (fully.west);
      \draw[->, thick] (fully.east) -- (output.west);

    \end{tikzpicture}